\documentclass[runningheads]{llncs}
\usepackage{graphicx}
\usepackage{mathrsfs}
\usepackage{subfigure}
\usepackage{balance}

\usepackage{array}
\usepackage{booktabs}
\usepackage{microtype}
\usepackage{url}
\usepackage{amsmath}
\usepackage{siunitx}
\usepackage{color}
\usepackage{multirow}
\usepackage{booktabs}
\usepackage{float}
\usepackage{bbding}
\usepackage{hyperref}

\usepackage{bm}
\usepackage{booktabs}
\usepackage{amssymb}% http://ctan.org/pkg/amssymb
\usepackage{colortbl}
\usepackage{multirow}
\usepackage{color}
\usepackage{arydshln}
\usepackage[utf8]{inputenc}

\usepackage{microtype}

\usepackage{inconsolata}

\definecolor{DarkGreen}{RGB}{0, 128, 0}
\definecolor{deltap}{RGB}{119, 139, 204}
\definecolor{deltan}{RGB}{255, 129, 90}

\usepackage{xspace}
\usepackage{url}

\usepackage[misc]{ifsym}
\usepackage{tabularx}
\usepackage{booktabs}
\usepackage{makecell}
\usepackage{multirow}
\usepackage{boldline}
\usepackage{caption}

\usepackage{wrapfig}
\usepackage{atbegshi}

\usepackage{algorithm}
\usepackage{algorithmic}

\usepackage{xcolor}

\begin{document}

\title{Boosting Neural Language Inference via Cascaded Interactive Reasoning}
\author{Min Li \and
Chun Yuan
}
\authorrunning{Li et al.}
\institute{Tsinghua Shenzhen International Graduate School, Tsinghua University, Beijing, China}

%\institute{Princeton University, Princeton NJ 08544, USA \and
%Springer Heidelberg, Tiergartenstr. 17, 69121 Heidelberg, Germany
\maketitle              % typeset the header of the contribution
\begin{abstract}
Natural Language Inference (NLI) focuses on ascertaining the logical relationship (entailment, contradiction, or neutral) between a given premise and hypothesis. This task presents significant challenges due to inherent linguistic features such as diverse phrasing, semantic complexity, and contextual nuances. While Pre-trained Language Models (PLMs) built upon the Transformer architecture have yielded substantial advancements in NLI, prevailing methods predominantly utilize representations from the terminal layer. This reliance on final-layer outputs may overlook valuable information encoded in intermediate layers, potentially limiting the capacity to model intricate semantic interactions effectively.
Addressing this gap, we introduce the Cascaded Interactive Reasoning Network (CIRN), a novel architecture designed for deeper semantic comprehension in NLI. CIRN implements a hierarchical feature extraction strategy across multiple network depths, operating within an interactive space where cross-sentence information is continuously integrated. This mechanism aims to mimic a process of progressive reasoning, transitioning from surface-level feature matching to uncovering more profound logical and semantic connections between the premise and hypothesis. By systematically mining latent semantic relationships at various representational levels, CIRN facilitates a more thorough understanding of the input pair. Comprehensive evaluations conducted on several standard NLI benchmark datasets reveal consistent performance gains achieved by CIRN over competitive baseline approaches, demonstrating the efficacy of leveraging multi-level interactive features for complex relational reasoning.

\keywords{
Neural language inference \and Deep learning \and Neural language process.
}
\end{abstract}
\section{Introduction}
\label{sec:intro}

Natural Language Inference (NLI), often framed as Recognizing Textual Entailment (RTE), seeks to discern the logical connection between a pair of sentences, classifying it as "entailment," "contradiction," or "neutral." As a foundational task in natural language understanding, NLI presents significant hurdles. Its difficulty arises not just from recognizing diverse linguistic constructions but also from the necessity of capturing underlying commonsense knowledge and performing nuanced reasoning.

The attention mechanism has proven highly effective across numerous NLP domains. It enhances cross-lingual semantic mapping in machine translation, aids in identifying crucial content for text summarization \cite{devlin2018bert}, and directs models towards relevant passages for accurate question answering. As formalized by \cite{vaswani2017attention}, attention calculates weighted context representations based on query-key similarities, adeptly modeling dependencies irrespective of distance in the sequence. Building on this success, we posit that developing more sophisticated attention-derived representations, which capture richer interaction patterns, can further elevate performance on complex reasoning tasks like NLI.

Conventional attention mechanisms typically compute an alignment matrix representing word-level correspondences between sentences. While multi-head attention \cite{vaswani2017attention} enriches this by offering multiple alignment perspectives, these often focus on pairwise token interactions. In this work, we aim to model more complex, higher-order dependencies that span beyond individual words to encompass phrase-level and structural relationships. We propose a method to generate intermediate representations encoding these intricate interactions, moving beyond simple lexical matching to capture structural configurations and combinatorial semantics between the sentence pair. Our experimental findings suggest that leveraging these enriched semantic features markedly improves logical reasoning accuracy and enhances model robustness in challenging scenarios involving paraphrases, antonyms, and lexical ambiguities.

To realize this, we introduce a novel architecture termed the Cascaded Interactive Reasoning Network (CIRN). CIRN is specifically designed to model intricate semantic interactions hierarchically. It operates by progressively extracting and integrating features across multiple layers of representation within an interactive space, simulating a gradual refinement of understanding from surface-level cues to deeper logical connections. We validate CIRN extensively on standard NLI benchmarks, including SNLI and MultiNLI \cite{bowman2015large}. Furthermore, to assess its versatility, we adapt CIRN for paraphrase identification, treating matched pairs as entailment and non-matching as neutral, and test it on the large-scale Quora Question Pair (QQP) dataset  and seven other datasets. Across these diverse evaluations, CIRN consistently surpasses strong baseline models, underscoring its effectiveness and broad applicability.

The primary contributions of this work can be outlined as follows:
\begin{itemize}
    \item Introduction of the Cascaded Interactive Reasoning Network (CIRN), a general framework designed for hierarchical aggregation and interactive reasoning over multi-level semantic information pertinent to NLI.
    \item A method embedded within CIRN to explicitly model complex, higher-order semantic dependencies and interactions between sentence pairs, extending beyond traditional token-level alignment approaches.
    \item Comprehensive empirical validation on multiple public datasets (including NLI and paraphrase identification), demonstrating consistent and significant performance improvements achieved by CIRN over competitive baseline methods.
\end{itemize}

\section{Related Work}
\label{sec:related_works}

\noindent \textbf{Traditional Neural Network-based Methods.}
Early approaches to Natural Language Inference (NLI) primarily relied on handcrafted syntactic features, transformation rules, and relation extraction techniques, which performed well on small-scale datasets but struggled with generalization and scalability~\cite{wang2007jeopardy}.
The introduction of large-scale datasets~\cite{bowman2015large} and the development of deep learning frameworks triggered a major shift in the field. Attention mechanisms, in particular, revolutionized NLI by enabling models to explicitly capture alignment and dependency relationships for enhanced semantic reasoning~\cite{conneau2017supervised,choi2018learning,liang2019adaptive}. Sentence-encoding architectures~\cite{conneau2017supervised} as well as joint models leveraging cross-attention mechanisms improved performance by learning hierarchical word- and phrase-level interactions~\cite{wang2017bilateral}. Innovations like residual networks further enhanced model depth while retaining lower-layer information~\cite{he2016deep,fei2022cqg}.
Neural network architectures have been widely applied for comprehensive sentence-level semantic modeling in NLI settings. Sequential dependencies and contextual semantic relationships are better modeled using Recurrent Neural Networks (RNNs)~\cite{liu2016learning,peng2020enhanced,wang-etal-2022-dabert,liu2023time}, while convolutional frameworks capture critical local features through sliding-window operations~\cite{xu2020enhanced}. Attention-based mechanisms further improve upon this by identifying salient components, aligning key phrases, and distilling semantic relevance~\cite{cho2015describing,xue2023dual,wu2025unleashing}. Techniques such as bidirectional Long Short-Term Memory (Bi-LSTM) networks offer a deeper representation of feature differences by incorporating contextual information at multiple levels~\cite{liu2023local,gui2018transferring}. Similarly, other architectures, including CNNs and RNNs, focus on either local sensitivity~\cite{yang2019simple} or sequential semantics~\cite{yang2019simple}. Recently, Graph Neural Networks (GNNs) have gained traction for leveraging sentence structural information and modeling global connections~\cite{dong2020distilling}. Additionally, some models adhere to the principle of encoding independent sentence representations without cross-interaction during feature extraction, relying on classifier layers for downstream tasks~\cite{choi2018learning}.

\begin{figure*}[ht]
\centering
\includegraphics[width=1.0\textwidth]{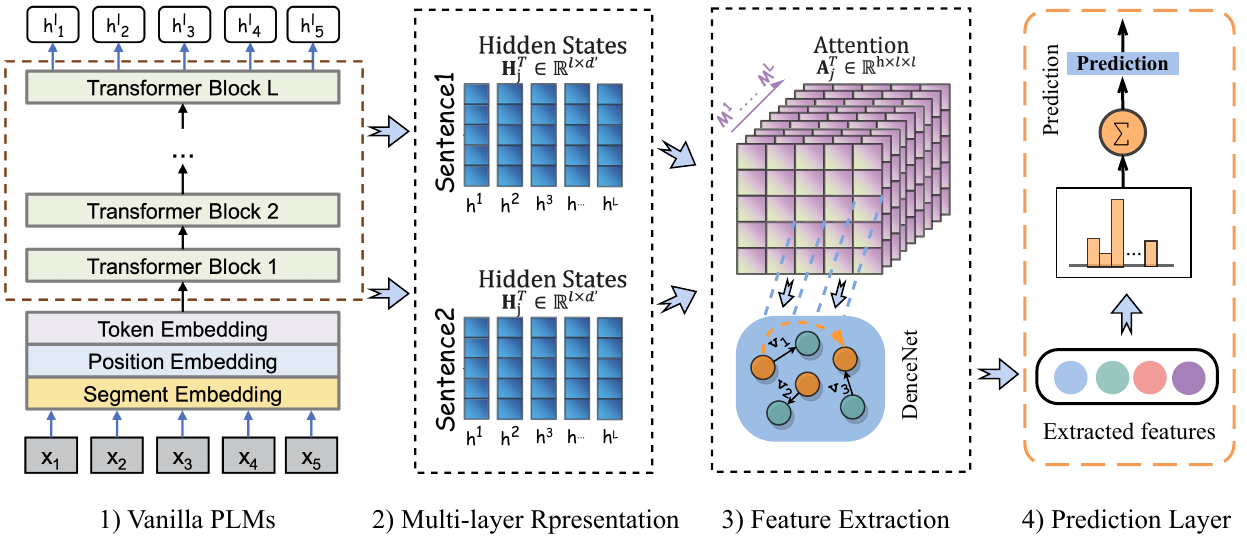}

\caption{
The overall architecture of the Cascaded Interactive Reasoning Network (CIRN): 1) \textbf{PLM Encoding}: concatenates sentence pair ($S_1$, $S_2$) as input to a pre-trained language model (e.g., BERT, RoBERTa), obtaining representations $\mathbf{H}^{(l)}$ from each Transformer layer; 2) \textbf{Multi-layer Representation}: separates $\mathbf{H}^{(l)}$ into sentence representations $\mathbf{H}_1^{(l)}$, $\mathbf{H}_2^{(l)}$, and computes interaction matrices $\mathbf{M}^{(l)} \in \mathbb{R}^{n \times m \times d}$ by element-wise multiplication; 3) \textbf{Feature Extraction}: stacks interaction matrices across layers into an interaction tensor $\overline{\mathbf{M}} \in \mathbb{R}^{n \times m \times d \times L}$, and applies DenseNet to extract deep interactive features; 4) \textbf{Prediction Layer}: aggregates extracted features through a fully connected layer followed by softmax to output classification probabilities.
}\label{fig:nli_model}
\end{figure*}

\noindent \textbf{Pre-trained Language Model-based Approaches.}
The emergence of pre-trained language models such as BERT~\cite{devlin2018bert} fundamentally transformed NLI tasks by introducing versatile representations obtained through pretext self-supervised objectives. These representations, fine-tuned on downstream tasks, led to significant gains in accuracy and robustness. Follow-up models such as XLNet~\cite{yang2019xlnet}, RoBERTa~\cite{liu2019roberta}, and CharBERT~\cite{xue2024question} further refined pretraining methodologies, thereby shrinking the performance gap between pretraining and fine-tuning. Cross-feature attention modules within these models allow them to explicitly focus on granular word- and phrase-level alignments, improving semantic reasoning~\cite{liang2019asynchronous,chen2016enhanced,ma2022searching,li2024comateformer}. Additionally, sentence interaction mechanisms, such as co-attentive DenseNets, further enhance multi-sentence relationships by integrating multiple layers of dependencies~\cite{song-etal-2022-improving-semantic,xue2024question}.
A complementary avenue of progress involves incorporating linguistic knowledge to refine sentence representations. Methods introduce explicit syntactic and semantic attributes, such as leveraging part-of-speech (POS) tags and named entities~\cite{li2024local,wu2024tablebench}, syntactic dependency parsing~\cite{liu2024resolving}, semantic roles~\cite{zhang2020semantics,chen2024s3prompt}, word synonym relations~\cite{xia2021using,zheng2022robust}, and syntax tree structures~\cite{bai2021syntax}. These augmentations help capture task-specific nuances, particularly for subtle distinctions that are not evident in global sentence semantics. However, current methods tend to prioritize global semantic similarity, often neglecting fine-grained distinctions at the token or phrase level, limiting performance on tasks requiring precise semantic matching.

\section{Model}
We frame the natural language inference (NLI) task as a multi-class classification problem aimed at predicting the relationship $y \in \mathcal{Y}$ between a sentence pair, where $\mathcal{Y} = \{\textit{entailment}, \textit{contradiction}, \textit{neutral}\}$. To tackle this, we propose the \textbf{Cascaded Interactive Reasoning Network (CIRN)}. The overall model architecture is depicted in the left panel of Figure \ref{fig:nli_model}.

\subsection{Input Preprocessing}
For a given sentence pair $S_1 = \{x_1^1, x_2^1, \dots, x_n^1\}$ and $S_2 = \{x_1^2, x_2^2, \dots, x_m^2\}$, we introduce special tokens [CLS] and [SEP] to conform to BERT's input format.
For BERT, the input embeddings are constructed as the sum of three components:
\begin{equation}
\label{eq:input_embedding}
\mathbf{X}_i = \mathbf{T}_i + \mathbf{S}_i + \mathbf{P}_i,
\end{equation}
where $\mathbf{T}_i$, $\mathbf{S}_i$, and $\mathbf{P}_i$ correspond to the Token Embedding, Segment Embedding, and Positional Embedding for the $i$-th token, respectively. Here, Segment Embedding differentiates the first text (Sentence A) from the second text (Sentence B).

\subsection{Multi-layer Transformer Representations}
The backbone of BERT is composed of $L$ stacked Transformer layers, each comprising Multi-Head Self-Attention and a Feed-Forward Network. Let $\mathbf{H}^{(0)} = [\mathbf{X}_1, \mathbf{X}_2, \dots, \mathbf{X}_{N'}]$ denote the input embedding sequence, where $N' = n + m + 3$ (including [CLS] and two [SEP] tokens). The output of the $l$-th Transformer layer can be expressed as:
\begin{equation}
\label{eq:transformer_layers}
\mathbf{H}^{(l)} = \text{TransformerLayer}_l\bigl(\mathbf{H}^{(l-1)}\bigr), \quad l = 1, 2, \dots, L.
\end{equation}
From $\mathbf{H}^{(l)}$, we extract the token embeddings for the first and second sentences, denoted as $\mathbf{H}_1^{(l)} \in \mathbb{R}^{n \times d}$ and $\mathbf{H}_2^{(l)} \in \mathbb{R}^{m \times d}$, where $d$ is the hidden dimension of the model.

\subsection{Interaction Matrices Across Layers}
\subsubsection{Element-wise Interaction Representation}
To explicitly model interactions between tokens from the two sentences, we compute an element-wise product between their embeddings. For the $l$-th Transformer layer, given token embeddings $\mathbf{h}_1^{(l, i)} \in \mathbb{R}^d$ (from the first sentence) and $\mathbf{h}_2^{(l, j)} \in \mathbb{R}^d$ (from the second sentence), the interaction tensor is defined as:
\begin{equation}
\label{eq:interaction_matrix}
\mathbf{I}^{(l)}_{i,j} = \mathbf{h}_1^{(l, i)} \odot \mathbf{h}_2^{(l, j)}, \quad \forall i \in [1, n], \, j \in [1, m],
\end{equation}
where $\odot$ represents element-wise multiplication. This operation produces a layer-specific interaction tensor $\mathbf{I}^{(l)} \in \mathbb{R}^{n \times m \times d}$, encoding token-level semantic dependencies for both sentences.

\subsubsection{Stacking Multi-layer Interactions}
The layer-wise interaction matrices are stacked along a new dimension to form a comprehensive multi-layer interaction representation:
\begin{equation}
\label{eq:multi_layer_stack}
\mathbf{I}_{\text{stack}} = \bigl[\mathbf{I}^{(1)}; \mathbf{I}^{(2)}; \dots; \mathbf{I}^{(L)}\bigr] \in \mathbb{R}^{n \times m \times d \times L},
\end{equation}
where $[\cdot;\cdot]$ denotes concatenation along the layer dimension. The aggregated tensor $\mathbf{I}_{\text{stack}}$ captures interactions at all layers, offering a detailed and hierarchical view of semantic relationships between the two sentences.

\subsection{DenseNet for Feature Extraction}
\label{sec:DenseNet}

DenseNet, originally designed for image processing, effectively enhances gradient flow and feature reuse through dense connections across layers. In our model, we adapt DenseNet to extract features from the stacked interaction tensor $\mathbf{I}_{\text{stack}}$. Convolutional or pooling operations are first applied to reduce dimensionality, followed by cascading dense blocks.

Let the input of the first dense block be $\mathbf{Z}_0$. Each layer in a dense block computes its output as:
\begin{equation}
\label{eq:densenet_layer}
\mathbf{z}_k = f_k\bigl(\text{Concat}(\mathbf{Z}_0, \mathbf{z}_1, \dots, \mathbf{z}_{k-1})\bigr),
\end{equation}
where $f_k$ represents the transformation function (e.g., convolution or fully connected operation) of layer $k$. The final output of the Dense Block is obtained as:
\begin{equation}
\label{eq:dense_block_output}
\mathbf{Z}_m = \text{Concat}(\mathbf{Z}_0, \mathbf{z}_1, \dots, \mathbf{z}_{m-1}),
\end{equation}
where $\mathbf{Z}_m$ aggregates features across all layers within the block.

\subsubsection{DenseNet Output Representation}
After passing $\mathbf{I}_{\text{stack}}$ through multiple Dense Blocks and Transition Layers, the final feature tensor $\mathbf{F}$ is derived:
\begin{equation}
\label{eq:final_features}
\mathbf{F} = \text{DenseNet}(\mathbf{I}_{\text{stack}}),
\end{equation}
where $\mathbf{F} \in \mathbb{R}^{d'}$ encapsulates high-level semantic interactions and refined sentence pair representations.

\subsection{Classification Layer}
The extracted features $\mathbf{F}$ are fed into a fully-connected layer, followed by a softmax classifier to predict the probability distribution over the relation categories $\mathcal{Y} = \{\textit{entailment}, \textit{contradiction}, \textit{neutral}\}$. Formally:
\begin{equation}
\label{eq:classification_layer}
\mathbf{p} = \text{softmax}\bigl(\mathbf{W}\mathbf{F} + \mathbf{b}\bigr),
\end{equation}
where $\mathbf{W} \in \mathbb{R}^{3 \times d'}, \mathbf{b} \in \mathbb{R}^3$ are trainable weights. The predicted label is selected as:
\begin{equation}
\label{eq:prediction}
\hat{y} = \arg \max_i \, \mathbf{p}_i.
\end{equation}

\subsubsection{Training Objective}
The training objective minimizes the cross-entropy loss between the true label $\mathbf{y}$ (one-hot encoded) and the predicted probability distribution $\mathbf{p}$:
\begin{equation}
\label{eq:cross_entropy_loss}
\mathcal{L} = - \sum_{i=1}^3 y_i \log(\mathbf{p}_i),
\end{equation}
where $\mathbf{y}$ represents the ground-truth label, and $y_i \in \{0, 1\}$ denotes the label for class $i$.

%\subsection{Overall Workflow}
%The CIRN model integrates preprocessing, interactive multi-layer representations, DenseNet feature extraction, and a classification layer. By explicitly modeling token-level and semantic dependencies across Transformer layers, it delivers richer feature representations and achieves state-of-the-art performance on natural language inference tasks.

\section{Experimental Configuration}
\label{sec:experimental_config}

This section details the experimental methodology used to evaluate our proposed Cascaded Interactive Reasoning Network (CIRN). We describe the datasets employed, the baseline models selected for comparison, and the specific implementation settings and hyperparameters used during training and evaluation.

\subsection{Evaluation Benchmarks}
\label{subsec:eval_datasets}
To thoroughly assess the performance of CIRN, our experiments utilize ten large-scale, publicly accessible benchmark datasets commonly employed in natural language understanding research.
A primary evaluation suite is the General Language Understanding Evaluation (GLUE) benchmark~\cite{wang2018glue}, a standard testbed featuring diverse NLP tasks including sentence similarity and textual entailment assessment\footnote{\url{https://huggingface.co/datasets/glue}}. We focus on six sentence-pair classification tasks within GLUE: MRPC, QQP, STS-B, MNLI, RTE, and QNLI.
Complementing the GLUE tasks, we incorporate four additional widely-used datasets to gauge model generalizability across different domains and task nuances: SNLI~\cite{bowman2015large}, SICK~\cite{wang2018glue}, TwitterURL~\cite{lan2017continuously}, and Scitail~\cite{khot2018scitail}.

\subsection{Comparative Baselines}
\label{subsec:comparative_baselines}

We benchmark CIRN against a representative set of existing models to demonstrate its relative effectiveness. Our main points of comparison are Pre-trained Language Models (PLMs), including the foundational BERT~\cite{devlin2018bert} architecture. We also compare against several PLM variants enhanced with external knowledge sources, namely SemBERT~\cite{zhang2020semantics}, SyntaxBERT~\cite{bai2021syntax}, and UERBERT~\cite{xia2021using}, allowing us to contextualize the gains from our specific interactive reasoning approach versus other knowledge integration techniques.
Furthermore, we include competitive models that do not rely on large-scale pre-training. This category features architectures like ESIM~\cite{chen2016enhanced}, a standard Transformer encoder~\cite{vaswani2017attention} trained specifically for the task, and other relevant models focusing on bilateral multi-perspective matching or compare-aggregate strategies~\cite{wang2017bilateral,tay2017compare}. These baselines help isolate the benefits derived from pre-training versus architectural innovations like CIRN.
For robustness evaluations (details potentially in another section referencing TextFlint~\cite{gui-etal-2022-kat}), performance comparisons are primarily made against the standard BERT baseline under various input perturbations. Detailed architectural descriptions of baseline models are omitted for conciseness and can be found in their respective publications.

\subsection{Implementation and Training Parameters}
\label{subsec:impl_details}

Our implementation utilizes the PyTorch deep learning framework. Optimization of all trainable model parameters is performed using the Adadelta optimizer~\cite{zeiler2012adadelta}, configured with $\rho = 0.95$ and $\epsilon = 1 \times 10^{-8}$. The initial learning rate is set to 0.5, and training proceeds with a batch size of 70. A learning rate adaptation mechanism is employed: if validation performance on the target task does not improve for 30,000 consecutive steps, the optimizer is switched to Stochastic Gradient Descent (SGD) with a reduced learning rate of $3 \times 10^{-4}$ to potentially refine the solution towards a better local minimum. Dropout layers are inserted before all linear transformation layers and following the word embedding layer to mitigate overfitting.
L2 weight decay is applied to all model weights for regularization. The effective L2 decay coefficient at training step $t$, denoted $\mathcal{R}_{L2}(t)$, is scheduled dynamically:
\begin{equation}
\label{eq:dynamic_l2_decay} % Renamed label
\mathcal{R}_{L2}(t) = \sigma\left(\frac{(t - T_{\text{ramp}} / 2) \times 8}{T_{\text{ramp}} / 2}\right) \times R_{L2}^{\text{max}}
\end{equation}
where $R_{L2}^{\text{max}}$ represents the maximum L2 decay ratio, set to $0.9 \times 10^{-5}$, and $T_{\text{ramp}}$ defines the number of steps over which the decay ramps up to its maximum, set to 100,000. $\sigma(\cdot)$ signifies the sigmoid function. Additionally, a specific L2 penalty with a coefficient of $1 \times 10^{-3}$ is applied to the squared difference between weights of corresponding encoder layers (if a dual-encoder setup is relevant).
Regarding the DenseNet-based feature extraction component (as described in Section \ref{sec:DenseNet}), architectural choices include setting the number of layers $n$ within each dense block to 8 and the growth rate $g$ to 20. The initial feature map reduction ratio $\eta$ is 0.3, and the transition layer compression factor $\theta$ is 0.5. To manage computational resources, maximum sequence lengths are enforced via truncation: 48 tokens for MultiNLI, 32 for SNLI, and 24 for QQP. Following established practice~\cite{liu2016learning}, the MultiNLI training incorporates 15\% of the SNLI dataset.

\begin{table*}[t]
\caption{The performance comparison of CIRN with other methods.}
\label{citation-guide-outsideGlue}
\centering
 \renewcommand\arraystretch{2.0}
% \scalebox{1}{
%\setlength{\tabcolsep}{2.8mm}{
%\resizebox{0.9\linewidth}{!}{
% \scalebox{1}{
\resizebox{\linewidth}{!}{
\begin{tabular}{lcccccccccccc}
\toprule
\textbf{Model} & \textbf{Pre-train}&\textbf{MRPC} & \textbf{QQP} & \textbf{STS-B} & \textbf{MNLI-m/mm}& \textbf{QNLI} & \textbf{RTE}  & \textbf{SNLI} & \textbf{Sci} & \textbf{SICK} & \textbf{Twi}  & \textbf{Avg}\\
\midrule
BiMPM &\XSolidBrush & 79.6 & 85.0 & - & 72.3/72.1 & 81.4 & 56.4 & - & - & - & - & -   \\
CAFE &\XSolidBrush & 82.4 & 88.0 & - & 78.7/77.9 & 81.5 & 56.8 & 88.5 & 83.3 & 72.3 & - & -  \\
ESIM &\XSolidBrush & 80.3 & 88.2 & - & 75.8/75.6 & 80.5 &  - & 88.0 & 70.6 & 71.8 & - & -   \\
% \hline
% \text{Transformer$\dagger$ \cite{vaswani2017attention}$\quad$} &\XSolidBrush& 81.7 & 84.4 & 70.4 & 72.3/71.4 & 80.3 & 58.1 & 81.7 & 70.6 & - & - & -\\
Transformer &\XSolidBrush& 81.7 & 84.4 & 73.6 & 72.3/71.4 & 80.3 & 58.0 & 84.6 & 72.9 & 70.3 & 68.8 & 74.4 \\
% \textbf{CAN$\dagger$$\quad$}  &\XSolidBrush& \textbf{82.1} & \textbf{84.9} & \textbf{72.6} & \textbf{72.6/72.5} & \textbf{81.3} & \textbf{59.2} & \textbf{84.2} & \textbf{76.4} & \textbf{-} & \textbf{-}& \textbf{-} \\
\hline
BiLSTM+ELMo+Attnt&\Checkmark & 84.6 & 86.7 & 73.3 & 76.4/76.1 & 79.8 & 56.8 & 89.0 & 85.8 & 78.9 & 81.4 & 78.9  \\
OpenAI GPT &\Checkmark & 82.3 & 81.3 & 80.0 & 82.1/81.4 & 87.4 & 56.0 & 88.4 & 84.8 & 79.5 & 81.9 & 80.4  \\
UERBERT &\Checkmark & 88.3 & 90.5 & 85.1 & 84.2/83.5 & 90.6 & 67.1 & 90.8 & 92.2 & 87.8 & 86.2 & 86.0   \\
SemBERT & \Checkmark & 88.2 & 90.2 & 87.3 & 84.4/84.0 & 90.9 & 69.3 & 90.9 & 92.5 & 87.9 & 86.8 & 86.5  \\
SyntaxBERT &\Checkmark  & 89.2 & 89.6 & 88.1 & 84.9/84.6 & 91.1 & 68.9  & 91.0 & 92.7 & 88.7 & 87.3  & 86.3\\
DABERT &\Checkmark  & 89.1 &  91.3 & 88.2 & 84.9/84.7 & 91.4 & 69.5 & 91.3 & 93.6 & 88.6 & 87.5 & 86.7\\
% DABERT &\Checkmark& 87.8 & 91.0 & 88.3 & 84.3/84.1 & 90.9 & 71.2 & 90.8 & 92.4 & 88.4 & 87.7 & 87.0 \\
%LG-SGC &\Checkmark & 89.5 & 91.3 & 88.6 & 85.5/85.4 & 92.3 & 73.9 & 92.1 & 94.0 & 90.1 & 89.3 & 88.3 \\
\hline
BERT-Base&\Checkmark & 87.2 & 89.1 & 86.8 & 84.3/83.7 & 90.4 & 67.2 & 90.7 & 91.8 & 87.2 & 84.8 & 85.8\\
BERT-Base-CIRN &\Checkmark & \textbf{89.3} & \textbf{89.1} & \textbf{87.1} & \textbf{85.1/84.9} & \textbf{91.2} & \textbf{68.5} & \textbf{91.2} & \textbf{92.1} & \textbf{87.8} & \textbf{86.6} & \textbf{86.8}\\
\hline
BERT-Large&\Checkmark & 88.9 & 89.3 & 87.6 & 86.8/86.3 & 92.7 & 70.1 & 91.0 & 94.4 & 91.1 & 91.5 & 88.0\\
BERT-Large-CIRN &\Checkmark & \textbf{89.9} & \textbf{90.2} & \textbf{88.1} & \textbf{86.8/86.7} & \textbf{93.0} & \textbf{72.0} & \textbf{91.1} & \textbf{94.3} & \textbf{91.2} & \textbf{92.4} & \textbf{88.8}\\
\hline
RoBERTa-Base &\Checkmark& 89.3 & 89.6 & 87.4 & 86.3/86.2 & 92.2 & 73.6 & 90.8 & 92.3 & 87.9 & 85.9 & 87.6 \\
RoBERTa-Base-CIRN  &\Checkmark & \textbf{89.5} & \textbf{91.2} & \textbf{88.1} & \textbf{87.7/87.5} & \textbf{93.2} & \textbf{82.5} & \textbf{91.2} & \textbf{93.1} & \textbf{89.5} & \textbf{87.6} & \textbf{89.2}\\
\hline
 RoBERTa-Large &\Checkmark& 89.4 & 89.7 & 90.2 & 89.5/89.3 & 92.7 & 83.8 & 91.2 & 94.3 & 91.2 & 91.9 & 90.3\\
 RoBERTa-Large-CIRN &\Checkmark  & \textbf{90.2} & \textbf{91.2} & \textbf{90.6} & \textbf{90.2/90.1} & \textbf{94.2} & \textbf{84.1} & \textbf{91.2} & \textbf{94.2} & \textbf{91.4}  & \textbf{92.3}& \textbf{90.7}\\
\bottomrule
\end{tabular}
}
%}

\end{table*}

\section{Results and Analysis} % Assuming this is the main section title
\subsection{Model Performance Evaluation} % Renamed subsection for clarity

Table \ref{citation-guide-outsideGlue} presents a performance comparison between CIRN~ and various competing approaches across 10 benchmark datasets. Consistent with expectations, methods based on pre-trained language models significantly surpass non-pretrained counterparts, owing to the extensive knowledge acquired during pre-training and sophisticated contextual representation capabilities.
Our proposed CIRN~ demonstrates notable improvements over strong baseline models. When integrated with BERT-base, it achieves an average accuracy gain of 0.8\%, and with BERT-large, the improvement is 0.7\%. These results highlight the efficacy of CIRN~ for tasks requiring nuanced language understanding, such as neural language inference. Furthermore, CIRN~ surpasses the performance of RoBERTa-base \cite{liu2019roberta} by 1.5\% and RoBERTa-large by 0.5\% on average across the evaluated datasets.
We attribute these performance enhancements to CIRN's effective modeling strategy, which concurrently analyzes sentence relationships from both fine-grained local interactions and broader global contexts. This multi-perspective approach enables the model to discern more subtle and intricate semantic connections that might be missed by methods focusing on only one level of analysis. The empirical evidence strongly suggests that incorporating mechanisms to model both local difference information and global semantic context, as CIRN~ does, is advantageous for semantic analysis tasks.
Overall, the experimental results position CIRN~ as a highly competitive method for evaluating semantic similarity and related tasks, empirically validating the effectiveness of our proposed architecture and approach across diverse benchmarks.

\begin{table*}
\centering
\renewcommand\arraystretch{1.5}

\small
\caption{Results of ablation experiment of various composition functions.\label{citation-guide-ablation}}
\setlength{\tabcolsep}{4mm}{

\begin{tabular}{l r r r}

\toprule
\multirow{2}*{\bf Ablation Experiments} & 
\multicolumn{2}{c}{\bf Dev Accuracy} \\

{}    & {\bf Matched} & {\bf Mismatched}\\
\midrule
1. CIRN$_{bert\_base}$ & 85.1 & 84.9 \\
% 3. DIIN - char feature & 78.6 & 78.6 \\
% 4. DIIN - POS feature & 78.6 & 78.2 \\
\midrule
2. CIRN$_{bert\_base}$ - Remove first 11 layer & 84.6 & 83.9 \\
3. CIRN$_{bert\_base}$ - Remove interaction matrix & 84.7 & 84.1 \\
4. CIRN$_{bert\_base}$ - Remove feature extraction & 83.9 & 83.4 \\
\bottomrule
\end{tabular}}

\end{table*}

\subsection{Ablation Study}
\label{subsec:ablation}

To isolate and evaluate the contribution of individual architectural components within our proposed model, we conducted an ablation study on the MultiNLI dataset \cite{williams2018broad}. The results of this analysis are presented in Table~\ref{citation-guide-ablation}.
First, we assessed the \textbf{impact of the interaction mechanism}. Removing the component responsible for element-wise interactions (specifically, the Hadamard product) resulted in a noticeable decrease in performance, with accuracy dropping from 85.1\% to 84.6\% on MNLI. This finding underscores the utility of this interaction feature for capturing fine-grained semantic relationships between the input sequences.
Second, we examined the \textbf{role of the DenseNet module}. Excising DenseNet from the architecture led to a more significant performance decline, reducing accuracy by 1.2\% on MNLI. This highlights DenseNet's crucial function in effectively extracting and consolidating high-level features and complex semantic patterns derived from the text interactions.
Third, the \textbf{advantages of multi-layer representation integration} were evaluated. When the model was restricted to using only the semantic information from the final layer of the underlying BERT encoder, performance degraded considerably compared to the full model which leverages representations from multiple layers. This result confirms the importance of integrating hierarchical semantic information captured across different depths of the Transformer architecture.
Collectively, these ablation experiments validate that the superior performance of our model is not attributable to a single component but arises from the synergistic contribution of its key elements. Both the detailed interaction modeling (via the interaction matrix) and the deep feature extraction capabilities facilitated by DenseNet, built upon multi-layer semantic inputs, are integral to the model's effectiveness.

\section{Conclusion}
\label{sec:conclusion}

In this paper, we delve into multi-level semantic feature extraction and propose a Cascaded Interactive Reasoning Network (CIRN). By aggregating multi-level semantic information, CIRN improves the accuracy of natural language reasoning, while CIRN breaks away from the limitations of traditional alignment. It encodes high-order complex relationships between sentence pairs, considering not only simple word-to-word connections, but also deeper lexical and semantic associations. This expands the receptive field of the model and enhances its ability to capture complex semantics.
The experimental results on 10 public datasets  show consistent improvements, especially the significant improvement in robustness tests, highlighting the effectiveness of CIRN.

\bibliographystyle{splncs04}
%\bibliography{mybibfile,anthology}

\end{document}